\documentclass[a4paper,fleqn,sort&compress]{cas-dc}

\usepackage{enumitem}
\usepackage{float}
\graphicspath{{./figs/}}
\usepackage{subfigure}
\usepackage{algorithmic}
\usepackage[linesnumbered,ruled]{algorithm2e}

\usepackage[numbers,sort&compress]{natbib}
\usepackage{amsmath}

\def\tsc#1{\csdef{#1}{\textsc{\lowercase{#1}}\xspace}}
\tsc{WGM}
\tsc{QE}

\begin{document}
\let\WriteBookmarks\relax
\def\floatpagepagefraction{1}
\def\textpagefraction{.001}

\shorttitle{arxiv}    
%
\shortauthors{Qing Tian et~al.}  
%
\title [mode = title]{Unsupervised Domain Adaptation Through Transferring both the Source-Knowledge and Target-Relatedness Simultaneously}  
%
%
%
\author[1,2]{Qing Tian}[style=chinese,]
\cormark[1]
\ead{tianqing@nuist.edu.cn}

\author[1,2]{Yanan Zhu}[style=chinese]
\ead{20201220059@nuist.edu.cn}

\author[1,2]{Chuang Ma}[style=chinese]
\ead{mcboo@nuist.edu.cn}

\author[3]{Meng Cao}[style=chinese]
\ead{alrash@nuist.edu.cn}

\affiliation[1]{organization={School of Computer and Software, Nanjing University of Information Science and Technology},
	 city={Nanjing,},
	 citysep={},
	 country={China}}

\affiliation[2]{organization={Engineering Research Center of Digital Forensics, Ministry of Education, Nanjing University of Information Science and Technology},
	city={Nanjing,},
	citysep={},
	country={China}}

\affiliation[3]{organization={College of Computer Science and Technology, Nanjing University of Aeronautics and Astronautic},
	city={Nanjing,},
	citysep={}, 
	country={China}}

%
%
%
%
%
%
%
%
%
%
%
%
%
%
%
%
%
\begin{abstract}
Unsupervised domain adaptation (UDA) is an emerging research topic in the field of machine learning and pattern recognition, which aims to help the learning of unlabeled target domain by transferring knowledge from the source domain. 
\end{abstract}



\begin{keywords}
	Multi-target domains\sep
	Source-knowledge\sep
	Target-relatedness
\end{keywords}

\maketitle


\section{Introduction}
\label{sec:introduction}

\begin{table*}[htbp]
	\centering
		\caption{Definition of symbols involved in this paper.\label{tab:notation}}
				\begin{tabular}{cc l}
					\toprule
					Notation & Dimension & {Meaning}\\
					\midrule
					$d$ & $\mathbb{R}$ & The feature dimension of the data samples  \\ 
					$K$, $K_T^m$ & $\mathbb{R}$, $\mathbb{R}$ & The classes number of the source domain and the $m$th target domain, respectively \\ 
					$N_S$, $N_T^m$ & $\mathbb{R}$, $\mathbb{R}$ & The samples number of the source domain and the $m$th target domain, respectively  \\ 
					$M$ & $\mathbb{R}$ & The number of target domains \\ 
					${{\textbf{W}}_S}$, ${\textbf{W}}_T^m$ & ${\mathbb{R}^{d \times K}}$, ${\mathbb{R}^{d \times K_T^m}}$ & The projection matrices of the source domain and the $m$th target domain, respectively \\ 
					${{\textbf{U}}^m}$& ${\mathbb{R}^{K_T^m \times N_T^m}}$ & The clustering membership matrix on the $m$th target domain  \\ 
					${\textbf{x}}_{T,i}^m$ & ${\mathbb{R}^d}$ & The $i$th instance from the $m$th target domain \\ 
					${\textbf{D}}$ & ${\mathbb{R}^{d \times r}}$ & The target-relatedness dictionary  \\ 
					${{\textbf{V}}^m}$ & ${\mathbb{R}^{K \times K_T^m}}$ & The transfer component matrix on the $m$th target domain from the source domain  \\ 
					${\textbf{V}}_T^m$ & ${\mathbb{R}^{r \times K_T^m}}$ & The target-relatedness component matrix on the $m$th target domain  \\ 
					${\textbf{Q}}$ & ${\mathbb{R}^{d \times d}}$ & The transfer transforming matrix \\
					\bottomrule
				\end{tabular}
\end{table*}

In this paper, we concentrate on 1SmT and proposes a UDA model through transferring both the Source-Knowledge and Target-Relatedness, coined as UDA-SKTR for short. In addition, we also present an alternating optimization algorithm to solve the proposed model with convergence guarantee. Overall, the contributions of this paper are three-fold as follows.
\begin{enumerate}[topsep=0em,itemsep=0em,parsep=0em]
	\item A 1SmT UDA model, coined as UDA-SKTR for short, is constructed by transferring both the source-knowledge and the target-relatedness, which is solved by a specially designed alternating algorithm with convergence guarantee.
	\item Different from existing methods that transfer directly from source domain data, we perform domain knowledge transfer from the source domain projection rather than the source domain data itself, which better protecting the privacy of the source data.
	\item Extensive evaluation experiments testify the effectiveness and superiority of the proposed method.
\end{enumerate}

\section{Related work} \label{sec:related work}
In this section, we briefly review several methods mostly related to our work, i.e., SLMC.

\subsection{SLMC}
\label{sec:SLMC}
Soft Large-Margin Clustering (SLMC) \cite{wang2013soft} is typical clustering method from the viewpoint of label space along the large-margin principle. It combines the advantages of soft label and large-margin clustering and achieves great performance in many scenes of clustering. For the objective function of SLMC, it is defined as follows:

\begin{align}
\label{eq1}
& \min_{\{\textbf{W}, u_{ki}\}} \frac{1}{2}\left\| {\textbf{W}} \right\|_F^2 + \frac{\lambda }{2}\sum\limits_{k = 1}^K {\sum\limits_{i = 1}^n {u_{ki}^2} } \left\| {{{\bf{W}}^T}{\textbf{x}_i} - {\textbf{l}_k}} \right\|_2^2\notag\\
& s.t.\quad u_{ki} \in [0,1],~ \sum_{k = 1}^K u_{ki} = 1
\end{align}

where $\textbf{W} \in \mathbb{R}^{d \times K}$ is the projection matrix with $d$ being the feature dimension and $K$ being the total number of data classes, $n$ means the number of instances, ${u_{ki}}$ denotes the clustering degree of membership for the $i$th instance to class $k$, and ${l_k} = {[0, \cdots ,0,1,0, \cdots ,0]^T} \in \mathbb{R}^K$ represents one-hot coding labels for the $k$th class with the $k$th element being 1 while other elements being 0.

\section{Unsupervised Domain Adaptation through transferring source-knowledge and target-relatedness}\label{sec:the proposed method}

\subsection{Formulation}\label{sec:Formulation}

\subsubsection{Knowledge transfer from source domain to target domains}\label{sec:source-knowledge}

To perform 1SmT UDA, we obviously need to transfer knowledge from the source domain to the target domains. Considering that the target domains are not consistent with each other and the domain shift from the source to these targets, it is necessary to introduce a transforming matrix (denoted as $\textbf{Q}$) to increase their matching flexibility. Along this line, we can mathematically formulate the scheme above as
\begin{align}\label{eq6}
	 \min_{\{\textbf{W}_T^m, \textbf{Q}, \textbf{V}^m\}} &\sum_{m = 1}^M {\left( {\left\| {\textbf{W}_T^m - \textbf{QW}_S{\textbf{V}^m}} \right\|_F^2 + \alpha {{\left\| {{\textbf{V}^m}} \right\|}_{2,1}}} \right)} \notag\\
	 s.t.~~& \textbf{Q}^T\textbf{Q} = \textbf{I}
\end{align}
In \eqref{eq6}, the first term is responsible for knowledge transfer from the source domain $\textbf{W}_S$ to the target domains $\textbf{W}_T^m$, characterized by domain transforming matrix $\textbf{Q}$ and domain representation matrix $\textbf{V}^m$. $\textbf{W}_S$ is learned from K-means algorithm. The second term encourages the UDA learning to select the most related knowledge components from the source domain to the targets. $\alpha$ is a nonnegative tradeoff parameter to keep a balance between the two terms. In order to prevent degenerated solutions of the transforming matrix $\textbf{Q}$, we restrict it column-orthogonal by $\bf{{Q^T}Q = I}$.

\subsubsection{Knowledge transfer between target domains}\label{sec:target-relatedness}
In the 1SmT scenario, multiple target domains are involved, which frequently exhibit potential promising correlations among them. Such relatedness may contribute to the target models training. To exploit these shared knowledge among these target domains, we propose to establish a over-complete representation dictionary to potentially extract such target-relatedness. Along this line, we can consequently construct the formulation for knowledge transfer between the multi-target domains as follows,
\begin{equation}\label{eq7}
	\scalebox{1}
	{$
		\begin{split}
			& \min_{\{{\textbf{W}}_T^m, \textbf{D},\textbf{V}_T^m\}} \; \sum_{m=1}^M \left\| {{\textbf{W}}_T^m - \textbf{DV}_T^m} \right\|_F^2 + \beta {\left\| {\textbf{V}_T^m} \right\|_{2,1}}
		\end{split}
		$}
\end{equation}

In \eqref{eq7}, the shared dictionary $\textbf{D}$ among the $M$ target domains plays the role of bridging them and exploring their relatedness to facilitate their learning. More specifically, the shared dictionary $\textbf{D}$ is established in the targets common space and it is over-complete to cover each of the target domains. That is, the projection matrix $\textbf{W}_T^m$ for the $m$th target domain can be recovered by $\textbf{D}$ with its reconstruction coefficient $\textbf{V}_T^m$. In this way, the potential relatedness among the target domains is integrated into their learning. The second term of \eqref{eq7} aims to select the most knowledge components from the dictionary to corresponding target domain.

\subsubsection{Unsupervised Domain adaptation by considering both source-knowledge and target-relatedness}
Through taking into account both the considerations aforementioned in Section \ref{sec:source-knowledge} and \ref{sec:target-relatedness}, we achieve incorporating both the source-knowledge and target-relatedness into 1SmT UDA. In consideration that we concentrate on supervised source domain and unsupervised target domains, without loss of generality, we readily take the SLMC objective function \eqref{eq1} for target domain clustering. Eventually, we can consequently build the complete objective function of the UDA model through transferring Source-Knowledge and Target-Relatedness, UDA-SKTR for short, as follows,
\begin{align}\label{eq8}
	& \min_{\{\textbf{W}_T^m,\textbf{D},\textbf{V}_T^m,\textbf{V}_S,u_{k,i}^m,\textbf{Q}\}} \; \sum_{m=1}^M \Bigg(\frac{1}{2}\sum_{k=1}^{K_T^m}\sum_{i=1}^{N_T^m}(u_{k,i}^m)^2\|\textbf{l}_k^m - (\textbf{W}_T^m)^T\textbf{x}_{T,i}^m\|^2  \notag\\
	&\qquad \qquad \qquad +\frac{\lambda_1}{2}\|\textbf{W}_T^m\|_F^2 + \frac{\lambda_2}{2}\|\textbf{W}_T^m - \textbf{QW}_S\textbf{V}^m\|_F^2 \notag\\
	& \qquad \qquad \qquad + \frac{\lambda_3}{2}\|\textbf{W}_T^m - \textbf{DV}_T^m\|_F^2 \Bigg)   \notag\\
	& \qquad \qquad \qquad + \lambda_4\Bigg(\|\textbf{V}_S\|_{2,1} + \sum_{m=1}^M\|\textbf{V}_T^m\|_{2,1}\Bigg)   \notag\\
	& \;\qquad s.t. \qquad \sum_{k=1}^{K_T^m}u_{k,i}^m = 1, \; 1 \leq m \leq M  \notag\\
	& \qquad \qquad \qquad 0 \leq u_{k,i}^m \leq 1  \notag\\
	& \qquad \qquad \qquad {\textbf{Q}^T}\textbf{Q} = \textbf{I}
\end{align}
where $\lambda_1$ to $\lambda_4$ are nonnegative tradeoff parameters. The first two terms are the objective w.r.t. SLMC on the target domains, the third term models target-relatedness among the $M$ target domains, while the fourth term transfers knowledge from the source to the target domains.

\subsection{Time complexity analysis}

The time complexity of UDA-SKTR  is mainly consisted of the alternating optimization steps.  Assume that our model converges after ${L^{\max}}$ iterations, and let ${N^{\max}}$ and ${K^{\max}}$ denote the maximum sample number and class number of the target domains. Taking into accounts all the time costs, we conclude that the total time complexity is ${\cal O}({LN_T^{\max}{d^2} + LN_T^{\max}d{{(K_T^{\max})}^3}{\rm{ + }}L{d^3}})$.

\newpage

\section{Conclusion}\label{sec:conclusion}

In this paper, we proposed a kind of 1SmT UDA model through transferring both the Source-Knowledge and Target-Relatedness, i.e., UDA-SKTR. In this way, not only the supervision knowledge from the source domain, but also the potential relatedness among the target domains are simultaneously modeled for exploitation in 1SmT UDA. In addition, we constructed an alternating optimization algorithm to solve the variables of the proposed model with convergence guarantee. Finally, through extensive experiments on both benchmark and real datasets, we validated the effectiveness and superiority of the proposed method. In the future, we will consider to extend the model to more challenging multi-source multi-target (mSmT) scenarios and extend it to practical application, such as  disease detection\cite{JIN2020106122} in medical field and fault detection\cite{JIAO2020106236} in industry.





%





\printcredits

\bibliographystyle{model1b-num-names}

\bibliography{manuscript}

%
%

%

\end{document}